\documentclass{new_tlp}
\usepackage{mathptmx}
\usepackage{xspace}
\usepackage{amsmath}
\usepackage{amssymb}
\usepackage{pifont}
\usepackage{hyperref}
\usepackage{listings}
\usepackage{csquotes}
\usepackage{todonotes}
\usepackage{graphicx}
\usepackage{caption}
\newcommand{\cmark}{\ding{51}}%
\newcommand{\xmark}{\ding{55}}%

\def\ar{\leftarrow}
\def\rar{\rightarrow}

\newcommand{\At}{\mathit{At}}

\newcommand{\hd}{\mathit{hd}}

\def\true{\mathit{true}}
\def\false{\mathit{false}}
\def\cB{\ensuremath{\mathcal{B}}}
\def \ez{{EZ}\xspace}

\def \ez{{EZ}\xspace}

\def \diff{{\sc cmodels(diff)}\xspace}
\def \ezsmt{{\sc ezsmt}\xspace}

\def \smtlib{{\sc smt-lib}\xspace}

\def \bbox{{\em blackbox}\xspace}

\def\bprolog{{\sc b}prolog\xspace}
\def\sp{{\sc swi}prolog\xspace}
\def\mzn{{\sc minizinc}\xspace}
\def\swi{{\sc swi}\xspace}
\def\mzna{{\sc mzn}\xspace}
\usepackage{epsfig}
\def\cvcFour{{\sc cvc4}\xspace}
\def\zThree{{\sc z3}\xspace}
\def\aspmt{{\sc aspmt2smt}\xspace}
\def\yices{{\sc yices}\xspace}

\def\cmodels{{\sc cmodels}\xspace}
\def \diff{{\sc cmodels(diff)}\xspace}
\def\clingo{{\sc clingo}\xspace}
\def\acsolver{{\sc acsolver}\xspace}
\def\clingodl{{\sc clingo[dl]}\xspace}
\def\clingolp{{\sc clingo[lp]}\xspace}

\def\gringo{{\sc gringo}\xspace}

\def\dingo{{\sc dingo}\xspace}

\def\mingo{{\sc mingo}\xspace}

\def\clasp{{\sc clasp}\xspace}

\usepackage{tikz}
\usetikzlibrary{decorations.markings}

\usetikzlibrary{shapes,arrows}
\tikzstyle{vertexblue}=[circle, fill=blue,draw, inner sep=0pt, minimum size=6pt]
\tikzstyle{vertex}=[circle, draw, inner sep=0pt, minimum size=6pt]
\newcommand{\vertex}{\node[vertex]}

\newcommand{\vertexblue}{\node[vertexblue]}

\tikzstyle{decision} = [diamond, draw, fill=blue!20, 
text width=4.5em, text badly centered, node distance=3cm, inner sep=0pt]
\tikzstyle{block} = [rectangle, draw, fill=blue!20, 
text width=5em, text centered, rounded corners, minimum height=4em]
\tikzstyle{line} = [draw, -latex']
\tikzstyle{cloud} = [draw, ellipse,fill=red!20, node distance=3cm,
minimum height=2em]

\def\cvcFour{{\sc cvc4}\xspace}
\def\zThree{{\sc z3}\xspace}
\def\aspmt{{\sc aspmt2smt}\xspace}
\def\dingo{{\sc dingo}\xspace}
\def\mingo{{\sc mingo}\xspace}
\def\ezsmt{{\sc ezsmt}\xspace}
\def\ezsmt{\ezsmt\xspace}

\def\ezcsp{{\sc ezcsp}\xspace}

\def\beq{\begin{equation}} 
\def\eeq#1{\label{#1}\end{equation}} 
\def\beq{\begin{equation}}
\def\eeq#1{\label{#1}\end{equation}}
\def\ba{\begin{array}}
\def\ea{\end{array}}

\newtheorem{theorem}{Theorem}
\newtheorem{definition}[theorem]{Definition}

\newtheorem{example}{Example}

\def\clasp{{\sc clasp}\xspace}

\def\clingcon{{\sc clingcon}\xspace}

\def\At{{\mathit{At}}}

\def\bi{\begin{itemize}}
\def\ei{\end{itemize}}

\def\cF{\ensuremath{\mathcal{F}}}

\def\bN{\mathbb{N}}

\newcommand{\belist}{\begin{list}{$\bullet$}{\topsep=1pt \parsep=0pt \itemsep=1pt}} \newcommand{\enlist}{\end{list}}

\def \ls#1{\todo[inline,color=purple!40]{#1}}

\def\personbox#1{%
\hbox to 0pt{\hss\color{black}#1\hss}}

\def\person#1#2#3{%
\noalign{%
\global\count1 #1
\global\advance\count1 
\global\count3 #2\relax
\global\advance\count3 -#1\relax}%
\multispan{\count1}{\hfill}&%
\multispan{\count3}{\strut\color[gray]{0.7}\leaders\vrule\hfill\mbox{}
\hskip0pt plus -.5fill\hskip0pt plus.5fill}\crcr}

\def\cmodels{{\sc cmodels}\xspace}
\def\gringo{{\sc gringo}\xspace}

\def\clasp{{\sc clasp}\xspace}

\def\At{{\mathit{At}}}

\newtheorem{example*}{Example}
\newtheorem{theorem*}{Theorem}
\newtheorem{proposition*}{Proposition}
\newtheorem{lemma*}{Lemma}

\def\bi{\begin{itemize}}
\def\ei{\end{itemize}}

\def\lr{\hbox{lr}\xspace}

\def\clingcon{{\sc clingcon}\xspace}
\def\ezcsp{{\sc ezcsp}\xspace}
\def \ezsmt{{\sc ezsmt}\xspace}

\def \ez{{EZ}\xspace}

\def \diff{{\sc cmodels(diff)}\xspace}

\def \smtlib{{\sc smt-lib}\xspace}

\def\sp{{\sc swi}prolog\xspace}
\def\mzn{{\sc minizinc}\xspace}
\def\swi{{\sc swi}\xspace}
\def\mzna{{\sc mzn}\xspace}

\def\cvcFour{{\sc cvc4}\xspace}
\def\zThree{{\sc z3}\xspace}
\def\aspmt{{\sc aspmt2smt}\xspace}
\def\yices{{\sc yices}\xspace}

\def\cmodels{{\sc cmodels}\xspace}
\def \diff{{\sc cmodels(diff)}\xspace}
\def\clingo{{\sc clingo}\xspace}

\def\gringo{{\sc gringo}\xspace}

\def\dingo{{\sc dingo}\xspace}

\def\mingo{{\sc mingo}\xspace}

\def\clasp{{\sc clasp}\xspace}

\title[Theory and Practice of Logic Programming]
{Constraint Answer Set Programming:\\ Integrational and\\ Translational (or SMT-based) \\Approaches
	}

\author[Yuliya Lierler]{YULIYA LIERLER\\University of Nebraska Omaha\\ \email{ylierler@unomaha.edu}}
\jdate{March 2003}
\pubyear{2003}
\pagerange{\pageref{firstpage}--\pageref{lastpage}}

\begin{document}
\label{firstpage}
\maketitle

\begin{abstract}
    Constraint answer set programming or CASP, for short, is a hybrid approach in automated reasoning putting together the advances of  distinct research areas such as answer set programming, constraint processing, and satisfiability modulo theories.  
    Constraint answer set programming
 demonstrates promising results, including
the development of a multitude of solvers: {\acsolver},
{\clingcon}, {\ezcsp}, 
{\sc idp}, {\sc inca}, {\sc
  dingo},  {\sc mingo}, \aspmt, {\sc clingo[l,dl]}, and~\ezsmt.  It opens new horizons for
 declarative programming applications such as solving complex train scheduling problems.
Systems designed to find solutions to constraint answer set programs can be grouped according to their construction into, what we call, {\em integrational or translational} approaches.
    The focus of this paper is an overview of  the key ingredients of the design of constraint answer set solvers drawing distinctions and parallels between integrational and translational  approaches. The paper also provides a glimpse at the kind of programs its users develop by utilizing a CASP encoding of Travelling Salesman problem for illustration. In addition, we place the CASP technology on the map among its automated reasoning peers
    as well as discuss future possibilities for the development of CASP.
    Under consideration in Theory and Practice of
    Logic Programming (TPLP).
\end{abstract}


\section{Introduction}
Knowledge representation and automated reasoning are  areas of Artificial Intelligence that pay especial attention to understanding and  automating  various aspects of reasoning. Such traditionally separate fields of AI
  as answer set programming~(ASP)~\cite{nie99,mar99,bre11},
  propositional satisfiability~(SAT)~\cite{gom08}, 
 constraint (logic) programming (CSP/CLP)~\cite{ros08,jm94} 
 are 
 representatives of model search in automated reasoning. 
 These methods
  have been successfully used   in a myriad of  scientific and industrial applications  including
 space shuttle control~\cite{bal01,bal05},
  scheduling \cite{RiccaGAMLIL12}, 
 planning \cite{kau92,Rintanen12}, 
 hardware  verification  \cite{BiereCCSZ03,PrasadBG05}, 
 adaptive Linux package configuration~\cite{geb11}, 
 systems biology \cite{GebserKSTV10}, bioinformatics~\cite{pal04,pal10}, software engineering~\cite{coh08,gar11,bra12}. 

Often the combination of algorithmic techniques stemming from distinct subfields of automated reasoning is necessary. 
 For instance, problems in software verification require reasoning combining  propositional logic  with formalizations that include, among others, theories of strings and  arrays. These observations led to studies
targeting the development of hybrid (multi-logic) computational methods that put together distinct solving approaches suitable for different logics.
This has led to the development of  {\em hybrid} approaches that combine
algorithms and systems from different AI subfields.  
Constraint logic programming~\cite{jm94},  satisfiability modulo theories~(SMT)~\cite{nie06,BarretSST08,BarTin-14},
HEX-programs~\cite{eit05a}, VI-programs~\cite{cal07},  constraint answer set programming~(CASP)~\cite{elk04,mel08,lier14} 
are all examples of this trend. 
Constraint answer set programming is the focus of this paper. 

Constraint answer set programming allows one to combine the best of two
  different automated reasoning worlds: 
(1) the non-monotonic modeling capabilities and SAT-like solving
technology of ASP; and (2) constraint processing techniques for 
effective reasoning over non-Boolean constructs.
CASP
 demonstrates promising results.
  For instance, research by~Balduccini  on the design of
 CASP language \textsc{ezcsp} and on the corresponding solver
 yields an elegant, declarative solution to a complex industrial scheduling
 problem~\cite{bal11}. Similarly, system \clingodl provides the basis for solving complex train scheduling problems~\cite{abe19}.
It is also due to note 
the development of many CASP solvers in the past decade:  {\acsolver}~\cite{mel08},
{\clingcon}~\cite{geb09}, {\ezcsp}~\cite{lier17a}, 
{\sc idp}~\cite{idp}, {\sc inca}~\cite{dre11a}, {\sc
  dingo}~\cite{jan11},  {\sc mingo}~\cite{liu12}, \aspmt~\cite{Bartholomew2014}, {\sc clingo[l,dl]}~\cite{jan17}, and~\ezsmt~\cite{sus16b,shen18a}.  It is fair to say that CASP formalism together with the multitude of supporting tools opens new horizons for
 declarative programming applications.

There are two main approaches in developing CASP systems/solvers, that is,  tools for processing programs in constraint answer set programming and enumerating their solutions.
The first one goes after systems that, while processing CAS programs, rely on combining \hbox{algorithms/solvers} employed in ASP and constraint processing~\cite{mel08,geb09,lier17a}. We call this approach {\em integrational}.
The second one transforms a CAS program into an SMT formula, whose models are in prespecified relation with answer sets of the original program~\cite{jan11,lee13,sus16b,lie17,shen18a}. As a result a problem of finding solutions to CASP is transformed into a problem of finding models of SMT formula.  We call this approach {\em translational}. 
The translational approach also includes two systems that translate CAS programs into other formalisms than SMT, namely, mixed integer programming, system \mingo~\cite{liu12}, and answer set programming, system {\sc aspartame}~\cite{ban15}. 

The focus of this paper is an overview of  the key ingredients of the integrational and translational  approaches towards construction of CASP systems. The paper starts with the  presentation of constraint answer set programming in use to showcase the paradigm. In particular, we present a CASP formulation of Traveling Salesman Problem benchmark alongside its ASP formulation. We then proceed towards defining  formal concepts of constraint answer set programming. The main part of the paper is devoted to describing details behind the integrational and translational approaches utilizing examples of two representatives of these methods --- systems \ezcsp and \ezsmt, respectively.
The paper also presents some experimental data together with an overarching comparison between the existing CASP systems in uniform terminological terms. We conclude with the discussion on future directions, opportunities, and challenges of the CASP subfield of automated reasoning. Before proceeding to the main topic of this paper we spend some time on placing CASP on the map of the automated reasoning subfield of artificial intelligence.

\subsection*{CASP and its Relatives}
 \begin{figure}
     \centering
 \begin{tabular}{ccccc}
      CASP&=&ASP&+&Constraints  \\
      SMT&=&SAT&+&Constraints  \\
      CLP&=&LP&+&Constraints  \\
 \end{tabular}
  \centering
     \caption{Paradigms' Content}
     \label{fig:relation}
 \end{figure}
 
 \begin{figure}
     \centering
 \begin{tabular}{cccc}
      &Programming/modeling language&~~
      &Automated reasoning (solver/system/compiler)\\
      ASP/CASP&\cmark&&\cmark\\
      SAT/SMT&&&\cmark\\
      LP/CLP&\cmark&&\cmark\\
 \end{tabular}
     \caption{Ingredients of Paradigms}
     \label{fig:dpp}
 \end{figure}

The question that comes to mind is what are the unique features of CASP in comparison to related formalisms, in particular,  satisfiability modulo theories, constraint logic programming, and answer set programming. 
Before drawing parallels between the fields, let us recall principal ingredients of declarative programming that CASP is a good representative of.
In declarative approach to programming no reference to an algorithm on how exactly to compute a solution is given. Rather a program provides a description/specification of what constitutes a solution. Automated reasoning techniques are then used to find a solution to provided specification. Thus,  declarative programming paradigm provides a programmer with two ingredients:
\begin{enumerate}
    \item Programming/modeling language to express requirements on a solution, and
\item     Automated reasoning method to find a solution.
\end{enumerate}
     
\paragraph{CASP vs SMT.}     Intuitive visualizations in Figures~\ref{fig:relation} and~\ref{fig:dpp} are of use\footnote{ In Figure~\ref{fig:relation} we understand word \textit{Constraints} as in constraint satisfaction.} when we compare CASP and SMT. 
     Figure~\ref{fig:relation} makes it clear that the key lies in relation between ASP and SAT. Lierler provides a detailed comparison of ASP and SAT~\citeyear{lier16a}. Here we reiterate the main thesis of that work:
\begin{quote}
    Answer set programming provides a declarative constraint programming language,
while SAT does not.
\end{quote}
The same claim is captured in Figure~\ref{fig:dpp}.
Both ASP/CASP and SAT/SMT pairs provide a solid platform for solving difficult combinatorial search problems. Automated reasoning tools behind these paradigms, called solvers,  share a lot in common. Yet, only ASP/CASP pair supplies its users with  programming/modeling language -- language of logic programs -- meant to express requirements on a solution using logic programs. The DIMACS and SMT-LIB standard formats of SAT and SMT solvers, respectively, provide a uniform front end to these systems,  but they are not meant for direct encoding of problems' specifications. 


\paragraph{CASP vs CLP} As Figure~\ref{fig:relation} suggests the key distinction between CASP and CLP lies in the difference of underlying paradigms of ASP and logic programming (LP). Marek and Truszczy\'nski draw a parallel between these two declarative programming paradigms~\citeyear{mar99}. To summarize, in original logic programming~\cite{kow88a}, called Prolog, a {\em single} intended model is assigned to a logic program. The SLD-resolution~\cite{kow74} is at the heart of control mechanism behind Prolog implementations. Together with a logic program, a Prolog system expects a query. This query is then evaluated by means of SLD-resolution and a given program against an intended model. In answer set programming, a {\em family} of intended models (possibly an empty one) is assigned to a logic program. Each member of this family forms a solution to a problem encoded by the program. Rules of a logic program formulate restrictions/constraints on solutions. A program is typically evaluated by means of a grounder-solver pair. A grounder is responsible for eliminating variables occurring in a logic program in favor of suitable object constants resulting in a propositional program. A solver -- a system in spirit of SAT solvers~\cite{lier16a} -- is responsible for computing  answer sets (solutions) of a program. Thus, even though LP and ASP share the basic language of logic programs, their  programming methodologies and underlying solving/control technologies are different.

\paragraph{CASP vs ASP.} The origin of CASP methods lies in attempts to tackle a challenge posed by  the \textit{grounding bottleneck} of ASP.
Sometimes  when a considered problem contains variables ranging over a large integer domain grounding required in pure ASP may result in a propositional program  of a prohibitive size.  CASP provides means to handle these variables within {\em Constraints} of the paradigm (see Figure~\ref{fig:relation}). There is also an additional benefit of the paradigm.
For example, some CASP dialects provide means to express constraints over real numbers whereas traditional ASP lacks this capacity. Thus, CASP offers novel modeling capabilities in comparison to these of pure ASP. 

\section{Constraint Answer Set Programming via Traveling Salesman Problem Formalization}
Before we dive into formal definitions, we present the formalization of a variant of the  Traveling Salesman Problem~\cite{tsp1,tsp2} in both answer set programming and constraint answer set programming (in the sequel, when we refer to this conjunction we  write \textit{(constraint) answer set programming} or (C)ASP). 
(Constraint) answer set programming provides a general purpose modeling language that
supports elaboration tolerant solutions for search problems.
We use the same notion of the search problem as Brewka et al.~\shortcite{bre11}. Quoting from their work, a {\em search problem}~$P$ consists of a set of instances with each
{\em instance}~$I$ assigned a finite set~$S_P(I)$ of solutions. 
In (constraint) answer set programming to solve a
search problem~$P$, we construct a program~$\Pi_P$ that captures
problem specifications so that when extended with facts~$D_I$
representing an instance~$I$ of the problem, the answer sets of
$\Pi_P\cup D_I$ are in one to one correspondence with members in~$S_P(I)$. In other words, answer sets 
describe all solutions of problem~$P$ for the instance
$I$. Thus, solving  a search problem is reduced to finding a uniform
encoding of its specifications by means of a logic program.

\begin{figure}
\footnotesize	
	\begin{tabular}{l|ll}
		
		An instance  
		with max cost 4&Solution 1&Solution 2\\
		\begin{minipage}[t]{.33\textwidth}
			\flushleft
			\begin{tikzpicture}[shorten >=1pt,node distance=3cm,on grid,auto]
			\tikzstyle{state}=[shape=circle,thick,draw,minimum size=1.5cm]
			\vertexblue (a1) at (2.7,3) [label=left:$a$] {};  
			\vertex (b1) at (4,3) [label=right:$b$] {};
			\vertex (d1) at (2.7,0) [label=left:$d$] {};
			\vertex (c1) at (4,0) [label=right:$c$] {};
			\path[-,draw,thick]
			(a1) edge node {1}  (b1)
			(b1) edge node {1}  (c1)
			(c1) edge node {1}  (d1)
			(d1) edge node {1}  (a1)
			(a1) edge node {2}  (c1)
			(b1) edge node {2}  (d1);
			
			
			\end{tikzpicture}
			
		\end{minipage}
		&
		\begin{minipage}[t]{.33\textwidth}
			\flushleft
			\begin{tikzpicture}[shorten >=1pt,node distance=3cm,on grid,auto]
			\tikzstyle{state}=[shape=circle,thick,draw,minimum size=1.5cm]
			
			\vertexblue (a1) at (2.7,3) [label=left:$a$] {};  
			\vertex (b1) at (4,3) [label=right:$b$] {};
			\vertex (d1) at (2.7,0) [label=left:$d$] {};
			\vertex (c1) at (4,0) [label=right:$c$] {};
			\path[->,draw,green,thick]
			(a1) edge   (b1)
			(b1) edge   (c1)
			(c1) edge   (d1)
			(d1) edge   (a1);
			
			
			\end{tikzpicture}
		\end{minipage}
		&
		\begin{minipage}[t]{.33\textwidth}
			
			\begin{tikzpicture}[shorten >=1pt,node distance=3cm,on grid,auto]
			\tikzstyle{state}=[shape=circle,thick,draw,minimum size=1.5cm]
			
			\vertexblue (a1) at (2.7,3.5) [label=left:$a$] {};  
			\vertex (b1) at (4,3.5) [label=right:$b$] {};
			\vertex (d1) at (2.7,0.5) [label=left:$d$] {};
			\vertex (c1) at (4,0.5) [label=right:$c$] {};
			\path[->,draw,green,thick]
			(a1) edge   (d1)
			(d1) edge   (c1)
			(c1) edge   (b1)
			(b1) edge   (a1);
			
			
			\end{tikzpicture}
			
		\end{minipage}
		\\
		{\em Encoded in ASP with facts: }&{\em Only instances of route/2 in}&\\
		& {\em answer set 1:} & {\em answer set 2:}\\
		{\tt city(a). ... city(d).}&&\\
		{\tt initial(a).}&{\tt route(a,b)}&{\tt route(a,d)}\\
		{\tt road(a,b). ... road(b,d).}&{\tt route(b,c)}&{\tt route(d,c)}\\
		{\tt cost(a,b,1). ... cost(b,d,2).}&{\tt route(c,d)}&{\tt route(c,b)}\\
		{\tt maxCost(4).}&{\tt route(d,a)}&{\tt route(b,a)}\\
	\end{tabular}
	\normalsize
	\caption{Sample TS Instance and Solutions\label{fig:ts}}		 		
\end{figure}
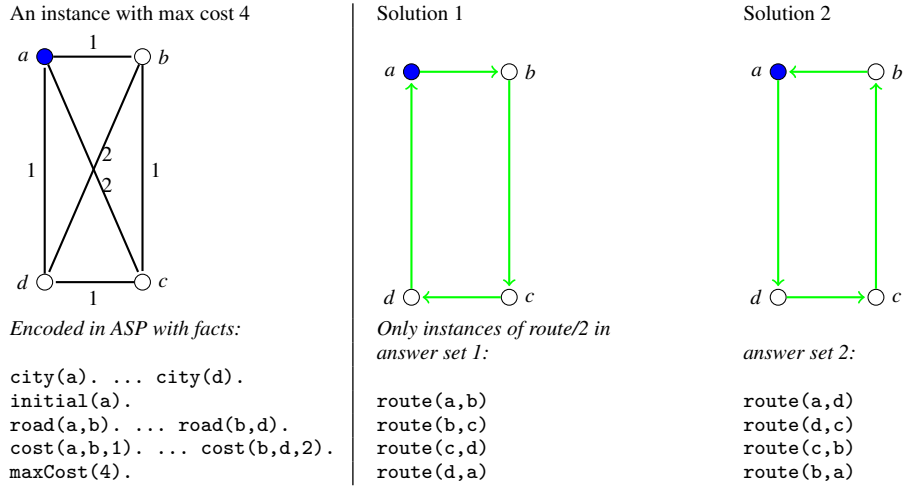
Consider the following combinatorial search problem: given an undirected weighted graph $G$ (where weights are non-negative integers), find a Hamiltonian cycle in $G$ with the sum of the weights of its edges at or below a given value. 
We can interpret  this problem as a {\em variant} of the {\em Traveling Salesman Problem} (TS):
\begin{displayquote}
{\em We are given a graph with  nodes as cities and edges as roads. Each road directly connects a pair of cities, and costs a salesman some time to go through (time is expressed as a positive integer value in this variant of the problem). The salesman is supposed to pass each city exactly once.
		 		Find:  {\em a route} traversing  {\em all the cities under} certain  {\em maximum cost} of total time. }
		 		\end{displayquote}
In the classical formulation of the TS problem, a route with the minimum cost  is of interest. 
Here we consider a decision problem in place of a related optimization problem. Also, in the classical formulation there are no restriction on weights over routes being integer.

Figure~\ref{fig:ts} shows an instance of the TS problem (a weighted graph) as well as its representation as a set of facts  (logic rules without bodies). 
On the right hand side  of the figure, we find two solutions to this problem.

Figure~\ref{fig:tsasp} presents an answer set programming formalization of the traveling salesman problem  using the syntax of the standard ASP-Core-2 Language~\cite{cal19}. 
Given a program composed of the rules in Figure~\ref{fig:tsasp} and the facts encoding the sample instance
 in Figure~\ref{fig:ts}, 
 an answer set solver such as \clingo, for example,  will produce the following output 
\begin{lstlisting}[
  basicstyle=\ttfamily\small
]
Answer: 1
route(a,d) route(c,b) route(d,c) route(b,a) 
Answer: 2
route(a,b) route(b,c) route(c,d) route(d,a) 
\end{lstlisting}
These answers correspond to the solutions of our sample instance.

\begin{figure}[h!]
	\small
				\begin{tabular}{l|l}
					\hline
					Encoding  & Meaning \\
					\hline
					{\tt road(Y,X):-road(X,Y).}& A road from X to Y is also a rode from Y to X.\\
					{\tt cost(Y,X,C):-cost(X,Y,C).}& A cost C for a road from X to Y  is also a cost\\
					&  for the same road from Y to X.\\
					&\\
					{\tt 1\{route(X,Y): road(X,Y)\}1:-city(X). }& For each city, pick one route {\em leaving from} the city. \\
					{\tt 1\{route(X,Y): road(X,Y)\}1:-city(Y). } &  For each city, pick one route {\em going to} the city. \\
					&\\
					{\tt reached(X):-initial(X).} &  The initial city is reached. \\
					 {\tt reached(Y):-reached(X), route(X,Y).} &  {If city X is reached and the route  from  X to  Y is }\\
					&  {picked, then city Y is also reached.} \\
					&\\
					{\tt :-city(X), not reached(X).} & A city that is not reached leads to a contradiction.\\
					&\\
					{\tt :-W$<$\#sum\{C,X,Y:route(X,Y),cost(X,Y,C)\}},
					  & {The total time cost of a selected route  greater than}\\
					  {\tt~~~~~~~~~~~~~~~~~~~~~~~~maxCost(W).}& maximal cost leads to a contradiction.\\
					  &\\
					  {\tt \#show route/2.}& A directive to only print route predicate as output.\\
					\hline
				\end{tabular}
\caption{TS: ASP encoding in the standard ASP-Core-2 Language\label{fig:tsasp}}
\normalsize
\end{figure}

\begin{figure}[h!]
		\includegraphics[scale=.75]{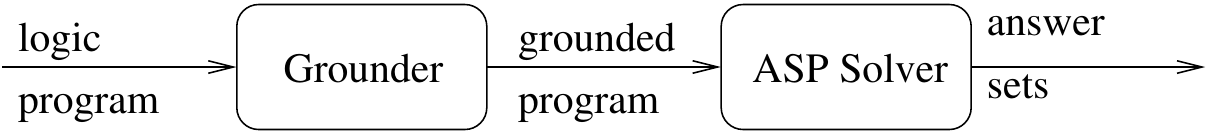}
	\caption{Answer Set Programming System Architecture\label{fig:aspsys}}
\end{figure}

\begin{figure}[h!]
	\small
				\begin{tabular}{l|l}
					\hline
				CASP Encoding& Meaning\\
					\hline
					
					
				{\tt cspvar(c(X,Y),0,C):-cost(X,Y,C).}  & Declaration of constraint variables.\\
				&\\
				{\tt  required(c(X,Y)==0):-cost(X,Y,C), not route(X,Y).}   & Time spent on a road is 0 if\\&
				road is not in  route.\\
					 {\tt required(c(X,Y)==C):-cost(X,Y,C), route(X,Y).}  &  Time spent on a road is its cost if \\& road is in route.\\
							 {\tt required(sum([c/2],$<$=,W)):- maxCost(W).}  &  Total time cost must be less or equal \\&to max cost.\\
					
					\hline
				\end{tabular}
				\caption{TS: Part of the CASP encoding in the \ez language of EZCSP \label{fig:tscasp}}
				\normalsize
			\end{figure}

Figure~\ref{fig:aspsys} presents a typical architecture of an answer set programming system. For example, aforementioned tool \clingo has this architecture. A grounder is a system that replaces non-ground rules (rules with variables) by their ground
counterparts (rules without variables/propositional rules)~\cite{geb07b,cal08}.  A solver is then invoked to find answer sets of a ground program. Procedures behind modern answer set solvers are close relatives of those behind SAT solvers~\cite{lier16a}. 
The process of grounding  in
ASP is well understood and highly optimized. For example, consider rule
\beq
\hbox{
	{\tt :-W$<$\#sum\{C,X,Y:route(X,Y),cost(X,Y,C)\}, maxCost(W).}}
\eeq{eq:rulesum}
from the ASP formalization of the TS problem and the discussed instance. A grounder of system {\clingo}
replaces  rule~\eqref{eq:rulesum}  with the following rule:

{\centering
{\tt 
:-4$<$\#sum\{1,a,b:route(a,b);1,b,c:route(b,c);1,c,d:route(c,d);\\
1,d,a:route(d,a);2,b,d:route(b,d);2,a,c:route(a,c);\\
1,b,a:route(b,a);1,c,b:route(c,b);1,d,c:route(d,c);\\
1,a,d:route(a,d);2,d,b:route(d,b);2,c,a:route(c,a)\}.
}

}
\noindent
 In some cases, the time taken by grounding dominates the time taken by solving. Addressing this difficulty is one of the challenges of ASP.

We now present the formulation of the TS problem using constraint answer set programming. 
In particular, we obtain a CASP encoding in the language of EZCSP by taking an ASP program given in Figure~\ref{fig:tsasp} and replacing its rule~\eqref{eq:rulesum} with lines presented in Figure~\ref{fig:tscasp}. In this encoding, we introduce \textit{constraint} variables $c(\cdot,\cdot)$ associated with each road so that when a road becomes a part of a route selected by a salesman its value is assigned to the cost of the road, while otherwise it is $0$. We then pose a constraint on these variables, which ensures that the total cost of a selected route is less than the maximal cost. 

The TS problem showcases some key  features that constraint answer set programming brings to the table in comparison to its parent -- answer set programming:
\begin{itemize}
\item Consider a simple change in the statement of the TS problem, namely,  
{\em time is expressed as a \textbf{real value}}. In fact, as mentioned earlier, the classical formulation of the TS problem considers weights that are real numbers. The traditional ASP framework may no longer be used to solve this problem. There is no support for real number arithmetic within grounders. Yet, CASP tools, such as, for example, \ezcsp or \ezsmt, can be used to find solutions to this new problem using the same program as presented here.
	\item 
ASP solvers process rules with so called sum-aggregates such as~\eqref{eq:rulesum} by implementing specialized procedures~\cite{nie00,geb09f,lierphd}.
By replacing~\eqref{eq:rulesum} with its CASP counterpart we
allow utilization of search techniques stemming from either
\begin{itemize}
 \item  CSP community  if we use such CASP tool as, for example, \ezcsp, or
\item  SMT community  if we use such CASP tool as, for example, \ezsmt.
 \end{itemize}
 These techniques will at times provide complementary performance.
 In other words, CASP allows us to utilize modeling language of ASP together with solving capabilities of SMT and CSP.
\end{itemize}
In addition,
\begin{itemize}
\item The grounding process of ASP may result in production of propositional programs that are of prohibitive size. This is especially the case when complex constraints over large numeric values are in place. CASP often allows us to bypass the grounding bottleneck via the reformulation of these numeric constraints using constraint atoms. Lierler~et al.~\citeyear{lierPadl12} presents a case study on  Weighted-Sequence problem (a domain inspired by a query optimization problem in relational databases), where the CASP solution is superior to its ASP counterpart as it alleviates grounding issues exhibited by an ASP solution.
\end{itemize}

 Just as a typical answer set solver, a common CASP system starts its computation by performing grounding on a given program.  
For example, such CASP systems as \ezcsp and \clingcon utilize grounder \gringo to produce a program composed of ground, so called, regular and irregular atoms.
For instance, consider a rule 
\beq
{\centering
\hbox{{\tt 
 required(c(X,Y)==C):- cost(X,Y,C),  route(X,Y).}
}}
\eeq{eq:route1}
\noindent
from the CASP TS encoding. We can view symbols $X$, $Y$, and $C$ as {\em schematic variables} that are placeholders for instances of passing constants.  
In the context of the CAS program composed of the sample TS instance in Figure~\ref{fig:ts} and the CASP TS encoding, rule~\eqref{eq:route1} will be grounded by the \ezsmt system into the  rules of the kind:

{\centering
{\tt 
required(c(a,b)==1):- cost(a,b,1),  route(a,b).\\
required(c(b,a)==1):- cost(b,a,1),  route(b,a).\\
$\cdots$\\
required(c(b,d)==2):- cost(b,d,2),  route(b,d).\\
required(c(d,b)==2):- cost(d,b,2),  route(d,b).\\
}
}

\noindent
As a result, a program that a solver component of a typical CASP system processes consists of 
\begin{itemize}
    \item [i.]  ``regular'' ground atoms such as $cost(a,b,1)$ and $route(a,b)$, and 
\item [ii.] ``irregular'' or constraint ground atoms such as $c(a,b)==1$ and $c(b,d)==2$, and 
\item[iii.] ground constraint variables such as $c(a,b)$ and $c(b,d)$.
\end{itemize}
Grounding process of CASP systems mirrors that of ASP systems. Thus, we direct a reader to papers by Gebser et. al~\shortcite{geb07b} and Calimeri et al.~\shortcite{cal08} for the details on grounding procedures. Here we focus on the unique features of CASP systems that pertain to their solving techniques. For this reason formal definitions that we present are in terms of ground/propositional CAS programs. 
We refer a reader, interested in the definition of syntax and semantics  for non-ground  CAS programs, to a paper by \citeN{bar13}.

\section{Preliminaries}\label{sec:intro} 

We now proceed towards formal preliminaries required to state the key definitions of the CASP paradigm.

\vspace{-3mm}
\paragraph{Logic Programs.}
A {\em vocabulary} is a set of propositional symbols also called atoms.
As customary, a {\em literal} is an atom~$a$ or its negation, denoted~$\neg a$.
A \emph{(propositional) logic program}, denoted by~$\Pi$, over vocabulary~$\sigma$  is a 
set of \emph{rules} of the form
\begin{equation}\label{e:rule}
\begin{array}{l}
a\ar b_1,\ldots, b_\ell,\ not\  b_{\ell+1},\ldots,\ not\  b_m,\ 
\ not\  \ not\  b_{m+1},\ldots,\ not\  \ not\  b_n,
\end{array}
\end{equation}
where $a$ is an atom over $\sigma$ or $\bot$, and each $b_i$, $1\leq i\leq n$, 
is an atom in $\sigma$.
We sometimes use the abbreviated form for  rule~\eqref{e:rule}
\begin{equation}\label{e:rulea:b}
\begin{array}{l}
a\ar B,\ 
\end{array}
\end{equation}
where $B$ stands for $b_1,\ldots, b_\ell,\ not\  b_{\ell+1},\ldots,\ not\  b_m,\ not\  \ not\  b_{m+1},\ldots,\ not\  \ not\  b_n$ and is also called a {\em body}. Syntactically, we identify rule~\eqref{e:rule} with the propositional formula
\begin{equation}\label{e:prop-formula}
b_1\wedge\ldots\wedge b_\ell\wedge \neg  b_{\ell+1} \wedge\ldots\wedge\neg  b_m \wedge \neg\neg b_{m+1} \wedge\ldots\wedge \neg\neg  b_n \rightarrow a
\end{equation}
and $B$ with the propositional formula
\begin{equation}\label{e:body-formula}
b_1\wedge\ldots\wedge b_\ell\wedge \neg  b_{\ell+1} \wedge\ldots\wedge\neg  b_m \wedge \neg\neg b_{m+1} \wedge\ldots\wedge \neg\neg  b_n.
\end{equation}
 Note (i) the order of terms in~\eqref{e:body-formula} is immaterial, (ii) {\emph{not} is replaced with classical negation ($\neg$)}, and (iii) comma is replaced with conjunction ($\wedge$).  
 Expression $$b_1\wedge\ldots\wedge b_\ell$$ in formula~\eqref{e:body-formula} is referred to as the {\em positive} part of the body and the remainder of~\eqref{e:body-formula} as the {\em negative} part of the body. Sometimes, we interpret semantically  rule~\eqref{e:rule} and its body as propositional formulas, in these cases it is obvious that double negation $\neg\neg$ in~\eqref{e:prop-formula} and~\eqref{e:body-formula}  can be dropped.


The expression $a$ is the \emph{head} of the rule. When $a$ is $\bot$, we often omit it and say that the head is empty. 
We call such rules {\em denials}.
We write $\hd(\Pi)$ for the set of nonempty heads of rules in~$\Pi$.
We call a rule whose body is empty a {\em fact}. In such cases, we drop the arrow. We sometimes may identify a set $X$ of atoms with the set of facts $\{a. \mid a \in X\}$.
For a logic program $\Pi$ (a propositional formula~$F$),
by $\At(\Pi)$ (by $\At(F)$) we denote the set of atoms occurring in $\Pi$ (in $F$). 

It is customary for a given vocabulary $\sigma$, to identify a set $X$ of atoms over $\sigma$ with (i) a complete and consistent set of literals over $\sigma$ constructed as $X\cup\{\neg a \mid a\in\sigma\setminus X\}$, and respectively with (ii)~an assignment function or interpretation that assigns truth value $\true$ to every atom in~$X$ and~$\false$ to every atom in $\sigma\setminus X$. 
We say a set~$X$ of atoms {\em satisfies} rule~\eqref{e:rule}, if~$X$ satisfies the propositional formula~\eqref{e:prop-formula}.
 We say~$X$ satisfies a program~$\Pi$, if~$X$ satisfies every rule in~$\Pi$. In this case, we also say that $X$ is a model of $\Pi$. We may denote satisfaction relation with symbol $\models$.

The {\sl reduct} $\Pi^X$ of a program $\Pi$ relative to a set $X$ of atoms is 
obtained by first removing all rules~\eqref{e:rule} such that $X$ does not satisfy negative part of the body
$$\neg  b_{\ell+1} \wedge\ldots\wedge\neg  b_m \wedge \neg\neg b_{m+1} \wedge\ldots\wedge \neg\neg  b_n,$$ 
and replacing all remaining rules with~$a\ar b_1,\ldots, b_\ell$ (note that $a$ can be $\bot$). 
\begin{definition}[Answer set]
	\label{def:answer-set}
A set~$X$ of atoms is an {\em answer~set}, if it is the minimal set that satisfies all rules of $\Pi^X$~\cite{lif99d}. 
\end{definition}

Ferraris and Lifschitz~\shortcite{fer05b} showed that a choice rule $\{a\} \ar B$ can be seen as an abbreviation for a rule $a \ar\ not\ not\ a, B$ (choice rules were introduced by Niemel\"a and Simons~\shortcite{nie00} and are commonly used in answer set programming languages). We adopt this abbreviation in the rest of the paper.

We now state the definition of an input answer set~\cite{lier11}
as it is instrumental in defining semantics for constraint answer set programs.
\vspace{-.6em}
\begin{definition}[Input answer set]
	\label{def:input-answer-set}
	For a logic program $\Pi$ over vocabulary~$\sigma$ and ({\em input/extensional}) vocabulary $\iota\subseteq\sigma$ such that none of $\iota$'s elements occur in the heads of rules in $\Pi$,
	a set~$X$ of atoms over~$\sigma$ is an \emph{input answer set} of~$\Pi$ relative to  $\iota$,  when $X$ is an answer set of the program
	$\Pi\cup (X\cap\iota)$.
\end{definition}

\begin{example}\label{ex:acp} Consider a logic program inspired by a running example  by \citeN{lier17a}:
\begin{equation}\label{eq:acp}
\ba l
  lightOn\ar\ switch,\ not\ am.\\
  \ar not\ lightOn. \\
\ea
\end{equation}
Take set $\{switch, am\}$ to form an input vocabulary. Intuitively,  a program is evaluated relative to truth values of these input atoms that are provided at the time of the evaluation.
Each rule in the program can be understood as follows:
\begin{itemize}
\item The light is on ({\em lightOn}) during the night ({\em not am}) when the action {\em switch} has occurred.
\item The light must be on.
\end{itemize}
Consider set $\{switch, \ lightOn\}$ of atoms. 
This set associates values $\true$ and $\false$ with input atoms $switch$ and $am$, respectively. 
This set is an input answer set of program~\eqref{eq:acp}.
Indeed, let $\Pi$ be  program~\eqref{eq:acp} extended with the fact $switch$. Reduct  $\Pi^{\{switch, \ lightOn\}}$ follows:
$$
\ba l
  switch. \\
  lightOn\ar\ switch.\\
\ea
$$
Set $\{switch, \ lightOn\}$ is an  answer set of this reduct. 
This set is the only input answer set of sample program~\eqref{eq:acp}.
This input answer set suggests that the only situation that satisfies the specifications of the problem is such that (i) it is currently night, (ii) the light has been switched on, and (iii) the light is on.
\end{example}
 
\paragraph{Input Completion.}
Clark~\shortcite{cla78} introduced the notion of program's completion. The process of completion turns a logic program into a classical logic formula. When a logic program satisfies certain  syntactic conditions,  models of a completion formula coincide with  answer sets of a logic program. In all cases, models of a completion formula include all answer sets of a logic program. Program's completion is a fundamental concept that plays an important role in the design of answer set solvers --- see, for instance, the paper by Lierler and Truszczynski~\shortcite{lier11}. It is also a major building block of the translational approach to   CASP solvers. We now review this concept together with the related notion of an input completion~\cite{lie17}.

Let $\Pi$ be a program over vocabulary~$\sigma$.
By $Bodies(\Pi,a)$ we denote the set of the bodies of all rules of~$\Pi$ with head~$a$.
The {\sl completion of program}~$\Pi$, denoted by $Comp(\Pi)$, is  the set of 
\begin{itemize}
\item classical formulas that consist of the rules~\eqref{e:rule}  in $\Pi$ (recall that we identify rule~\eqref{e:rule} with  implication~\eqref{e:prop-formula};
when a rule~\eqref{e:rule} is a fact $~a$, then we identify this rule  with the  clause consisting of a single atom~$a$)  and 
\item the implications
\beq
a\rar \bigvee_{a\ar B\in \Pi} B
\eeq{eq:comp2}
for all atoms $a$ in $\sigma$. When the set $Bodies(\Pi,a)$ is empty, the implication~\eqref{eq:comp2} has the form $a\rar\bot$.
\end{itemize}

We now define an input completion that is relative to an (input) vocabulary.
\begin{definition} [Input completion]
For a program $\Pi$ over vocabulary $\sigma$,
the {\em input-completion} of $\Pi$ relative to vocabulary~\hbox{$\iota\subseteq\sigma$} so that $\hd(\Pi)\cap\iota=\emptyset$, denoted by~$IComp(\Pi,\iota)$, is defined as the set of formulas in propositional logic that consists of the  rules~\eqref{e:prop-formula} in $\Pi$ and the implications~\eqref{eq:comp2}
for all atoms~$a$ occurring in $\sigma\setminus\iota$.
\end{definition}

\paragraph{Level Ranking.}
Niemel\"a~\citeyear{nie08} characterized answer sets of ``normal'' logic programs  in terms of program's completion and ``level ranking''.
Normal programs  consist of rules of the form~\eqref{e:rule}, where $n=m$ and $a$ is an atom. 
Lierler and Susman~\citeyear{lie17} generalized
a concept of a level ranking  to  programs introduced here.
These results are  fundamental in realizations of many translational approaches to (constraint) answer set programming. 
For instance,
Niemel\"a  developed a mapping from normal programs to the satisfiability modulo difference logic formalism (to be introduced in detail shortly).
That translation  paved the way towards the implementation of answer set solvers {\sc lp2diff}~\cite{jan09} and {\sc cmodels-diff}~\cite{shen18}.
Similarly,
translational constraint answer set solvers~{\mingo}~\cite{liu12}, {\dingo}~\cite{jan11}, {\sc aspartame}~\cite{ban15}, {\sc ezsmt}~\cite{shen18a}  rely on the concepts of completion and  level ranking (and its variants, i.e, strong level ranking and strongly connected component level ranking proposed by Niemel\"a) in devising their translations.

We start by introducing some notation to formally define the concept of level ranking that accommodates the notion of an input vocabulary. 
By  $\mathbb{N}$ we denote the set of natural numbers.
For a rule~\eqref{e:rulea:b}, by $B^+$ we denote  its  positive part  and sometimes identify it with the set of atoms that occur in it, i.e., $\{b_1,\dots, b_l\}$ (recall that $B$ in~\eqref{e:rulea:b} stands for the right hand side of the arrow in rule~\eqref{e:rule}).


 
 \begin{definition} [Level ranking]\label{def:lrinput}
  A function $\lr: X\setminus\iota \rightarrow \bN$ is a {\em level ranking} of $X$ for $\Pi$ relative to vocabulary $\iota\subseteq\sigma$ so that \hbox{$\hd(\Pi)\cap\iota=\emptyset$},  when  for every atom $a$ in $X\setminus\iota$ the following condition holds: 
  there is $B$ in $Bodies(\Pi,a)$ such that~$X$ satisfies $B$  and 
   	 for every $b \in B^+\setminus\iota$ it holds that $\text{\lr}(a) - 1 \geq \text{lr}(b)$.
  \end{definition}

 We now restate Theorem~8 from Lierler and Susman~\citeyear{lie17} that captures the relation between input answer sets of a program and models of input completion by means of level ranking.
 \begin{theorem}\label{thm:casp-ans-iff-lr}
	\label{thm:slr}
	For a program $\Pi$ over vocabulary $\sigma$, vocabulary $\iota\subseteq\sigma$ so that $\hd(\Pi)\cap\iota=\emptyset$, and a set~$X$ of atoms over $\sigma$ that is a model of input completion $IComp(\Pi,\iota)$, $X$ 
 is an input answer set of $\Pi$ relative to $\iota$ if and only if  there is a level ranking of $X$ for $\Pi$ relative to $\iota$.
\end{theorem}
This result is related to the characterization of answer sets of a logic program as models of its completion~\cite{fag94}.

\paragraph{Constraints.}
Lierler and Susman~\citeyear{lie17} illustrated that the notion of a  ``constraint'' (as understood in classical literature on constraint processing within the artificial intelligence realm) coincides with the notion of a ground literal of satisfiability modulo theories. Furthermore, a  {\em constraint satisfaction problem (CSP)}, which is usually defined by a set of constraints, can be identified with the conjunction of ground literals. This conjunction is evaluated by means of first-order logic interpretations/structures  representative of a particular ``uniform'' SMT-theory -- a term introduced by Lierler and Susman~\shortcite{lie17}.
An {\em SMT-theory}~\cite{BarTin-14} is a set of interpretations/structures. 
A {\em uniform SMT-theory}~\cite{lie17} is  a set of interpretations whose domain, interpretation of predicates and ``interpreted" function symbols are fixed.

In practice, special forms of  constraints are commonly used. {\em Integer linear constraints} are examples of these special cases. For instance, 
\beq
2 x+ 3 y>0
\eeq{eq:scon}
is a common abbreviation for an integer   linear  constraint. In line with Lierler and Susman, we identify linear integer inequality~\eqref{eq:scon} with a   ground atom
$$>(+(\times(2,x),\times(3,y)),0),$$
where we assume an SMT-theory called {\em Integer Linear Arithmetic or Linear Integer Arithmetic (ILA)} (see, for instance, the paper by Bromberger et al.~\shortcite{bro15}). This theory is defined by the set of all possible interpretations,  whose  domain is the set of integers, the predicate $>$ is interpreted as an  arithmetic greater  relation/predicate symbol; function symbols $+$ and $\times$ are interpreted as usual in arithmetic; 0-arity function symbols $2$, $3$, and $0$ are
interpreted by mapping these into respective domain elements (identified with the same symbol).  The constraint~\eqref{eq:scon}
contains uninterpreted  0-arity function symbols $x$ and $y$ that are frequently referred to as object constants  (in logic literature)  or variables (in constraint processing literature). 

We call an interpretation satisfying a CSP, which we understand as the conjunction of ground literals,  its {\em solutions}. We identify this interpretation with a function called {\em valuation} that provides a mapping for uninterpreted function symbols to domain elements. For example, one of the solutions to the CSP composed of a single constraint~\eqref{eq:scon}   within ILA-theory is a valuation  that maps $x$ to $0$ and $y$ to $1$.
Formulas composed of integer linear constraints and interpreted using SMT-theory ILA are said to be within {\em ILA-logic}~\cite{BarTin-14}.

Other commonly used SMT-theories are called {\em difference logic} (DL)~\cite{nie05} and {\em linear arithmetic} (LA)~\cite{BarTin-14}. 
In difference logic the set of interpretation defining this theory is that of ILA. Yet, difference logic restricts the syntactic form of constraints to the following $x-y\leq k$, where $x$ and~$y$ are variables and $k$ is 0-arity function symbol interpreted by a mapping to domain elements (integers). Linear arithmetic logic differs from ILA-logic in its SMT-theory:
domain of linear arithmetic logic is a set of {\em real} numbers.

\section{Constraint Answer Set Programs and SMT Formulas, Formally}\label{sec:cassmt}
Let $\sigma_r$, $\sigma_e$, and $\sigma_i$ be three disjoint vocabularies.
We refer to their elements as \emph{regular}, \emph{strict-irregular} atoms, and  \emph{non-strict-irregular} atoms, respectively. The terms strict and non-strict are due to Gebser et al.~\shortcite{geb16}, where the authors introduce the CASP language that permits capturing two commonly used semantics in CASP dialects. 


\begin{definition}[Constraint answer set program and its answer sets]\label{def:casProgram}
	Let $\sigma=\sigma_r\cup \sigma_e
	\cup\sigma_i$ be a vocabulary so that regular atoms $\sigma_r$, strict-irregular atoms  $\sigma_e$, and non-strict-irregular atoms $\sigma_i$ 
 are disjoint; $\cB$ be a set of constraints; $\gamma$ be an injective
	function from the set of irregular literals over~$\sigma_e\cup \sigma_i$ to  $\cB$; and
 $\Pi$ be a logic
program over $\sigma$ such that $\hd(\Pi)\cap(\sigma_e\cup\sigma_i)=
\emptyset$.	
	We call a triple \hbox{$P=\langle \Pi,\cB,\gamma\rangle$} a {\em constraint answer set 
program} (CAS program)
	over vocabulary $\sigma$.

	A set $X\subseteq\At(\Pi)$ of atoms
	is an \emph{answer set}
	of $P$ if
	\begin{enumerate}
		\item[(a)]  $X$ is an input answer set of $\Pi$ relative to $\sigma_e\cup\sigma_i$, and
		\item[(b)] the  following CSP           has a solution:
		$$\{\gamma(a) \mid a\in X\cap(\sigma_e\cup \sigma_i)\}\cup \{\gamma(\neg a) \mid a\in \sigma_e\setminus X\}.
		$$
	\end{enumerate}
	A pair $\langle X,\nu\rangle$
	is an \emph{extended answer set}
	of $P$ if $X$ is an \emph{answer set} of $P$ and valuation $\nu$ is a solution to the CSP constructed in (b).
\end{definition}
It is now time to remark on the differences between  regular, {strict-irregular}, and  {non-strict-irregular} atoms.
If vocabulary $\sigma$ only consists of regular atoms  $\sigma_r$ (sets  $\sigma_e$ and
$\sigma_i$ of irregular atoms are empty) then CAS program turns into a logic program under answer set semantics. 
Per condition (a) all irregular atoms are part of the input/extensional vocabulary.
Intuitively, irregular atoms carry additional information that goes beyond their truth value assignment. This fact culminates in the statement of the (b) condition in the definition of an answer set. The (b) condition also points at the difference between strict-irregular and non-strict-irregular atoms. While the presence of irregular atoms in set $X$ of atoms requires a constraint of this atom to be satisfied, only the absence of a strict-irregular atom requires a constraint of its complement to be satisfied. The non-strict irregular atoms do not pose the latter restriction.

In the sequel, we  utilize vertical bars to mark irregular atoms that  have intuitive mappings into respective 
constraints. For instance, given an integer variable $x$, the expression $|x<0|$ corresponds to an irregular atom that is mapped into
constraint/inequality \hbox{$x<0$}; similarly
irregular literal  $\neg |x<0|$  is mapped into
constraint/inequality \hbox{$x\geq 0$}. 

\begin{example}\label{ex:casp} Let us  consider  CAS  program $P_1=\langle \Pi_1, \cB_1, \gamma_1\rangle$ 
from Example~3 by \citeN{lie17}. Logic program $\Pi_1$ --- the first element of the tuple defining $P_1$  --- follows
\begin{equation}\label{eq:casp}
\ba l
  \{switch\}.\\
  lightOn\ar\ switch,\ not\ am.\\
  \ar not\ lightOn. \\
  \{am\}.\\
  \ar not\ am, |x<12|.\\
  \ar am, |x \geq 12|.\\
\ar |x<0|.\\
\ar |x>23|.\\
\ea
\end{equation}
The set $\sigma_r$ of regular atoms of~$P_1$ is $$\{switch, am, lightOn\}.$$
The set $\sigma_e$ of strict-irregular atoms of~$P_1$ is
\beq
\{|x<0|,\; |x<12|,\; |x\geq12|,\; |x>23|\},
\eeq{ex:vocabir}
where $x$ is an integer variable (representing hours of the day). 
The set $\sigma_i$ of non-strict-irregular atoms of~$P_1$ is~empty.

The first line of the program is understood as follows: {\em The action {\em switch} is exogenous.}
The second two lines are identical to these of logic program \eqref{eq:acp}.
The fourth line we can intuitively read as: 
{\em It is night ({\em not am}) or morning ({\em am})}.
The last four lines of the program state:
\begin{itemize}
\item It must be $am$ when $x<12$.
\item It is impossible for it to be $am$ when $x \geq 12$.
\item Variable $x$ must be nonnegative.
\item Variable $x$ must be less than or equal to $23$.
\end{itemize}
\bigskip
Set  
$\cB_1$ consists of integer linear constraints   including constraints 
$$\{x<0,\; x\geq 0,\; x<12,\; x\geq 12,\;  x>23,\;  x\leq 23\},$$ 
	Mapping $\gamma_1$ is defined as follows
	 $$\gamma_1(a) = \begin{cases}
	 	\mbox{constraint $x<0$ }  &\mbox{if } a = |x<0| \\
				\mbox{constraint $x\geq 0$}  &\mbox{if } a = \neg |x < 0|\\
		\mbox{constraint $x<12$ }  &\mbox{if } a = |x<12|  \hbox{ or } a = \neg |x \geq 12|\\
		\mbox{constraint $x\geq 12$}  &\mbox{if } a = |x \geq 12| \hbox{ or } a = \neg |x < 12|\\
				\mbox{constraint $x> 23$}  &\mbox{if } a = |x > 23|\\
				\mbox{constraint $x\leq 23$}  &\mbox{if } a = \neg |x > 23|.
	\end{cases}$$

\noindent
Consider  set 
\begin{equation}\label{eq:caspAnswerSet}
	\{switch, \ lightOn, |x \geq 12|\}
\end{equation}
over the vocabulary of $P_1$. This set  is the only input answer set of $\Pi_1$ relative to irregular atoms of $P_1$. 
Also, the 
 integer linear constraint satisfaction problem
formed by the  constraints in 
$$
 \ba{c}
 \{
   \gamma_1(\neg |x < 0|),\;
 \gamma_1(\neg |x < 12|),\;
 \gamma_1(|x \geq 12|), \;
  \gamma_1(\neg |x > 23|)
 \}\\
 =\\
 \{ 
 x\geq 0,\;
 x\geq12,\;
 x\leq 23 
 \}
 \ea
 $$ has a solution. There are 12 valuations ${v_1}\dots v_{12}$ for integer variable $x$, which satisfy this CSP, namely, $x^{v_1}=12,\dots,x^{v_{12}}=23$. 
It follows that
set~\eqref{eq:caspAnswerSet} is an answer set of $P_1$.
Pair 	$$\langle \{switch, \ lightOn,\; |x \geq 12|\},\nu_1\rangle$$ 
is one of the twelve extended answer sets of $P_1$.

To illustrate the difference between strict and non-strict irregular atoms consider the CAS program~$P'_1$ that differs from $P_1$ only in sets $\sigma_e$ and $\sigma_i$.
In particular, the set $\sigma_e$ of strict-irregular atoms of~$P'_1$ is empty.
The set $\sigma_i$ of non-strict-irregular atoms of~$P'_1$ is~\eqref{ex:vocabir}.
Set~\eqref{eq:caspAnswerSet}
 is the only input answer set of $\Pi_1$ relative to irregular atoms of $P'_1$. 
Also, the 
 integer linear constraint satisfaction problem formed by the 
 constraint in $$
 \ba{c}
 \{\gamma_1(|x \geq 12|) 
 \}\\
 =\\
 \{x\geq12
 \}
 \ea
 $$ has a solution. There are indeed infinite number of  valuations ${v_1}\dots v_{12},\; v_{13}, \dots$ for integer variable $x$, which satisfy this CSP, namely, $x^{v_1}=12,\dots,x^{v_{12}}=23,\; x^{v_{13}}=24, \dots$.

 We direct a reader to the paper by Gebser et al.~\shortcite{geb16}, where the authors discuss in detail the rationale behind the two distinct kinds of irregular atoms.

\end{example}

We note that if we consider a CAS program  whose set $\sigma_i$ of non-strict-irregular atoms is empty then it falls into a class of programs accepted by such CASP system as \clingcon~\cite{geb09},  given that constraints are in the realm of ILA. Similarly, if we consider a CAS program  whose set $\sigma_e$ of strict-irregular atoms is empty 
and whose atoms from $\sigma_i$ only occur in denials,
then it falls into a class of programs that such  CASP system as \ezcsp accepts (given that constraints are in the realm of ILA or LA).

Janhunen et al.~\citeyear{jan17} lift the restriction on irregular atoms not to occur in the heads of program's rules. Rather, they  divide all atoms into ``defined''  and ``external'' (input/extensional, if to follow the terminology of this paper), where defined atoms may occur in heads. In other words, defined irregular atoms do not longer need to be part of the input vocabulary.  This is an important and an interesting extension within CASP that is utilized in the implementation of such CASP systems as \clingodl and \clingolp. To the best of our knowledge these are the only two currently available CASP systems that allow ``defined'' irregular atoms. 

\paragraph{SMT Formulas.}
Here we state the definition of an SMT formula~\cite{BarTin-14}. This concept is fundamental for most translational approaches to CASP.

\begin{definition}[SMT formulas and its models]
Let $\sigma=\sigma_r\cup \sigma_e
	\cup\sigma_i$ be a vocabulary (so that $\sigma_r$,  $\sigma_e$, and $\sigma_i$ 
 are disjoint); $\cB$ be a set of constraints; $\gamma$ be an injective
	function from the set of irregular literals over~$\sigma_e\cup \sigma_i$ to  $\cB$;
 $F$ be a propositional formula  over $\sigma$.	 
	We call a triple~$\cF=\langle F,\cB,\gamma\rangle$ an {\em SMT formula}
	over vocabulary~$\sigma$.
	A
	set \hbox{$X\subseteq\At(F)$} is a    {\em model} of SMT formula $\cF$
	if
	\begin{enumerate}		\item[(a.1)]  $X$ is a model of $F$, and
		\item[(b.1)] the CSP constructed in~(b) of Definition~\ref{def:casProgram} has a solution.
	\end{enumerate}
	A pair $\langle X,\nu\rangle$
	is an \emph{extended model}
	of $\cF$ if $X$ a \emph{model} of $\cF$ and $\nu$ is a solution to the CSP in~(b.1).
\end{definition}

We are now ready to provide the translation  from logic programs to SMT formulas (this translation is inspired by level ranking results). We  then  present a translation by Lierler and Susman~\shortcite{lie17} that maps CAS programs into SMT formulas.

As before, we  utilize vertical bars to mark irregular atoms (introduced within the translation) that  have intuitive mappings into respective 
constraints. For instance, the expression $|lr_b-1\geq lr_a|$ corresponds to an irregular atom that is mapped into
constraint/inequality \hbox{$lr_b-1\geq lr_a$}, where $lr_a$ and $lr_b$ are variables over integers. To refer to the constraints corresponding to  irregular literals we use superscript $\downarrow$. For example, 
$$
\ba{rll}
|lr_b-1\geq lr_a|^\downarrow&=&lr_b-1\geq lr_a\\ 
\neg |lr_b-1\geq lr_a|^\downarrow&=&lr_b-1< lr_a.
\ea
$$
\begin{definition}[Translation from a logic program to an SMT formula]\label{def:tr1}
 Let $\Pi$ be a logic program over vocabulary $\sigma^\Pi$ and vocabulary $\iota\subseteq\sigma^\Pi$ such that none of $\iota$'s  elements occur in the heads of rules in $\Pi$.
 For every atom $a$ in $\sigma\setminus\iota$ that occurs in~$\Pi$ we introduce an integer variable $lr_a$. The SMT formula $\cF^\Pi=\langle F^\Pi,\cB^\Pi,\gamma^\Pi\rangle$ is constructed as follows
 \begin{itemize}
\item  formula $F^\Pi$ is a conjunction of the following
\begin{enumerate}
        \item\label{ib1}  rules~\eqref{e:rule}  in $\Pi$;
        \item\label{ib2}
        for each atom $a\in\sigma^\Pi\setminus\iota$  the implication
        $~~~~
        a\rar
        \displaystyle\bigvee_{a\ar B\in \Pi }\Big(B\wedge
        \bigwedge_{b\in B^+\setminus\iota}   
        |lr_a-1\geq lr_{b}|
        \Big)
        $
\end{enumerate}
\item set $\sigma_r$ of $\cF^\Pi$ is formed by the atoms in $\sigma^\Pi$;
set $\sigma_e$ of $\cF^\Pi$ is formed by the
irregular atoms of the form $|lr_a-1\geq lr_{b}|$ introduced in~\ref{ib2}; and
set $\sigma_i$ of $\cF^\Pi$ is empty;
\item constraints in $\cB^\Pi$ are composed of inequalities
 $|lr_a-1\geq lr_{b}|^\downarrow$ and 
 $\neg |lr_a-1\geq lr_{b}|^\downarrow$
 for all irregular atoms of the form $|lr_a-1\geq lr_{b}|$ introduced in~\ref{ib2}; \item  function $\gamma^\Pi$ maps 
irregular literals formed from atoms of the form $|lr_a-1\geq lr_{b}|$ introduced in~\ref{ib2} 
to  constraints in $\cB^\Pi$ in a natural way captured by $\downarrow$ function.
\end{itemize}
\end{definition}
An SMT formula $\cF^\Pi$ has two properties: (i) it has models if and only if respective answer set program $\Pi$ has answer sets and (ii) any model $I$ of this formula is such that $I\cap\sigma$ forms an answer set of $\Pi$. This is a consequence of Theorem~9 by Lierler and Susman~\shortcite{lie17}, which follows from Theorem~\ref{thm:casp-ans-iff-lr} restated here.

\begin{definition}[Translation from a CAS program to an SMT formula]\label{def:tr2}
Let $P=\langle \Pi,\cB,\gamma\rangle$ be a CAS program over   $\sigma^P=\sigma_r^P\cup\sigma_i^P\cup\sigma_e^P$.
For every atom $a$ in $\sigma_r^P$ that occurs in~$\Pi$ we introduce an integer variable $lr_a$. The SMT formula $\cF^P=\langle F^P,\cB^P,\gamma^P\rangle$ over 
$\sigma=\sigma_r\cup\sigma_i\cup\sigma_e$
is constructed as follows 
\begin{itemize}
\item the formula  $F^P$ is a conjunction consisting of formulas in~\ref{ib1} and~\ref{ib2} of Definition~\ref{def:tr1},  where we understand $\sigma^\Pi$ as $\sigma^P$ and
$\iota$ as $\sigma_i^P\cup\sigma_e^P$; 
\item set $\sigma_r$ of $\cF^P$ is formed by the atoms in $\sigma_r^P$;
set $\sigma_e$ of $\cF^P$ is formed as the union of $\sigma_e^P$ and the
irregular atoms of the form $|lr_a-1\geq lr_{b}|$ 
described in Definition~\ref{def:tr1};
set $\sigma_i$ of $\cF^P$ is  formed by the atoms in $\sigma_i^P$;
\item constraints $\cB^P$ are composed of the elements in 
$\cB$ and the elements in $\cB^\Pi$ described in Definition~\ref{def:tr1};
\item mappings of $\gamma^P$ are composed of the elements in $\gamma$ and the elements in $\gamma^\Pi$  described in Definition~\ref{def:tr1}.
\end{itemize}
\end{definition}
An SMT formula $\cF^P$  has two properties: (i) it has models if and only if respective CAS program has answer sets and (ii) any model $I$ of this formula is such that $I\cap\sigma^P$ forms an answer set of $P$. This is a consequence of Theorem~10 by Lierler and  Susman~\shortcite{lie17} that follows from Theorem~\ref{thm:casp-ans-iff-lr} restated here.

\section{Integrational Approach via System \ezcsp}\label{sec:int}
As stated in the introduction, this paper presents the details behind two CASP systems, namely, \ezcsp\footnote{Solver \ezcsp is available at 
\url{http://mbal.tk/ezcsp/} .} and \ezsmt\footnote{\label{foot:ezsmt}Solver \ezsmt  is available at\\  \url{https://www.unomaha.edu/college-of-information-science-and-technology/natural-language-processing-and-knowledge-representation-lab/software/ezsmt.php} .}. The former is a  representative of the integrational approach. The latter is a representative of a translational SMT-based approach. 

Both systems, \ezcsp 
and \ezsmt,
accept programs written in the language that is best documented  by Balduccini and Lierler~\shortcite{lier17a}. We call this language \ez.
The TS problem formulation of this paper is in that language. 
This paragraph uses system \ezcsp in its claims. The same claims are applicable for the system \ezsmt.
As discussed earlier, CASP systems typically start their computation by grounding a given program. System \ezcsp uses grounder \gringo~\cite{Gebser2011} for this purpose.  
Following our example from the end of the introduction, ground rule 	
\beq \hbox{{\tt  required(c(a,b)==1):- cost(a,b,1),  route(a,b).}  }
\eeq{eq:sreq}
exemplifies the kinds of ground rules produced by the \ezcsp system at the time of grounding.
Atoms of the form  $required(\beta)$ instruct the \ezcsp system that~$\beta$ introduces a non-strict-irregular atom.
Even though  ``required atoms'' occur in the head of  rules, semantically these rules are denials with the irregular atom ``complementary'' to $\beta$ occurring in the body.\footnote{It is due to note that $\beta$ may be a more complex expression than an irregular atom. For example, it may contain a disjunction of irregular atoms. Semantically, a rule with such $\beta$ expression in its head corresponds to the denial that extends the body of this rule with the conjunction of the complementary atoms formed from $\beta$.} For instance,
rule~\eqref{eq:sreq} stands for the following denial,
written	in style used in Section~\ref{sec:cassmt}:
$$
\ar	cost(a,b,1),\ route(a,b),\ |c(a,b)\neq 1|.
$$ 
Atoms such as $|c(a,b)\neq 1|$ belong to non-strict-irregular atoms of the CAS program produced by grounding of \ezcsp. These CAS programs  (i) contain no strict-irregular atoms and (ii) contain irregular atoms only in denials. 

\begin{figure}[htbp]
\begin{center}
\includegraphics[clip=true,trim=0 170 0 0,width=1\columnwidth]{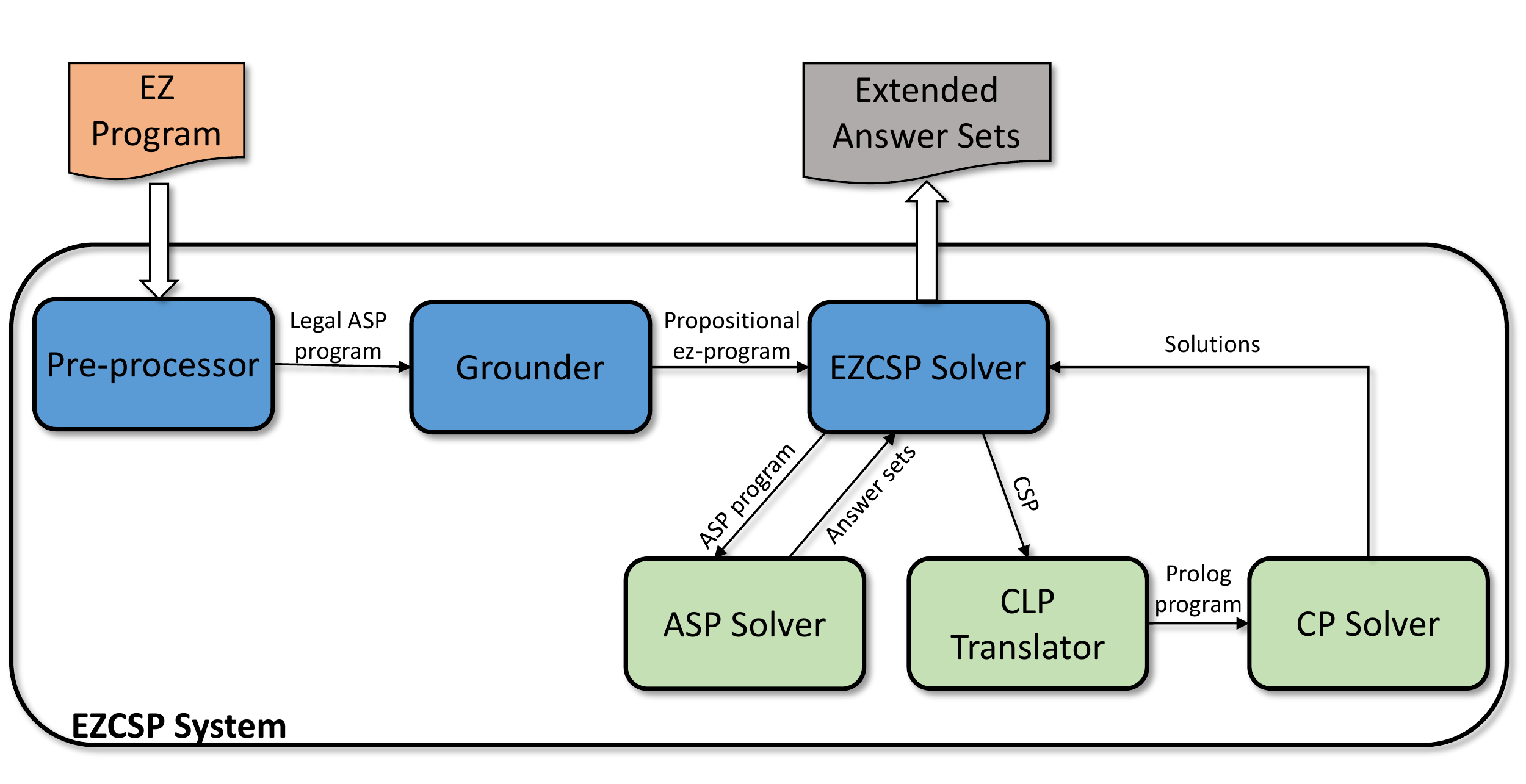}
\caption{Architecture of the {\ezcsp} system}\label{fig:arch}
\end{center}
\end{figure}

Figure~\ref{fig:arch} depicts the architecture of the \ezcsp system. The graphic is reproduced  from the paper by Balduccini and Lierler~\shortcite{lier17a}. 
We follow the presentation by Balduccini and Lierler
to state the most essential details behind the \ezcsp system. 
The first step of the execution of \ezcsp (corresponding to the \emph{Pre-processor} component in the figure) consists in running a {pre-processor, which} 
translates an input~{\ez} program  into a syntactically legal ASP program.
This is accomplished by replacing the occurrences of arithmetic functions and operators
in expressions of the form $required(\beta)$ by auxiliary function symbols. For example, an expression $v>2$ in
$required(v>2)$ is replaced by $gt(v, 2)$.
The \emph{Grounder} component of the architecture transforms the
resulting program into its propositional equivalent, a regular
program, using an off-the-shelf {grounder} such as
{\gringo}~\cite{geb07b,Gebser2011}. 
This regular program is then passed to the {\emph{{\ezcsp} Solver} component}.

The \emph{{\ezcsp} Solver} component iterates between ASP and constraint
programming computations by invoking the corresponding components of
the architecture. Specifically, the \emph{ASP\ Solver} component
computes an answer set of a given regular program using an off-the-shelf 
{ASP solver}, such as {\cmodels} or {\clasp}.
If an answer set is found, the {\ezcsp} solver runs the \emph{CLP
  Translator} component, which maps the CSP problem corresponding
to the computed answer set 
to a {Prolog program}. The program is then passed to the
\emph{CP\ Solver} component, which uses the {CLP tools} such as SICStus~\cite{sicstus-prolog}, SWI Prolog~\cite{swi-prolog} or {\bprolog}~\cite{b-prolog}, to solve
the CSP instance. Recent version of \ezcsp augment the \emph{CLP
  Translator} component with the possibility of producing MiniZinc~\cite{minizinc} formulations of  CSP problems. As a result, MiniZinc solvers\footnote{\url{https://www.minizinc.org/} .} can be used in place of CLP tools.
Finally, the \emph{{\ezcsp} Solver} component gathers the solutions to the respective CSP problem and combines
them with the answer set obtained earlier to form extended answer sets. Additional extended
answer sets are computed iteratively by finding other
answer sets and the solutions to the corresponding CSP problems.

It is essential to note that in integrational approaches the communication schemas between the two participating solving mechanisms are important. 
The presented architecture of \ezcsp showcases the so called  \bbox integration approach.
The beauty of the \bbox approach is its flexibility in utilizing the existing technology as both answer set solver and CSP solver can be taken as they are. Yet, it is obvious that such an integration does not provide any means to rely on advances in search of an answer set solver in earlier iterations or prune the computation of an answer set solver based on information from a CSP.  The \ezcsp system also implements so called \emph{grey-box and clear-box} integration, where it accommodates continuation in search and  early pruning, respectively. In this capacity the \ezcsp is confined to utilizing a particular answer set solver \cmodels via its internal API. 

\paragraph{Briefing: Integrational Systems.}
This is a good place to speak of some other integrational systems. We first consider solver \clingcon~\cite{geb09}. 
From the original design of the system to its latest version, its authors were proponents of a clear-box integration. Its original implementation established the clear-box communication between answer set solver \clasp (a solver of answer set system {\sc clingo})   and constraint processing system {\sc gecode}. The second, recent, implementation of \clingcon~\cite{ost17} uses  sophisticated  ``in house CSP'' propagators to replace the  {\sc gecode} system. It is also a close relative to
the newest representatives of the integrational approach
systems \clingodl and \clingolp~\cite{jan17}. Latest version of \clingcon and these two systems are a product of a systematic effort by  University of Potsdam  to create an extensible infrastructure to support answer set programming based solutions. Framework \clingo~5~\cite{geb16} provides comprehensive interfaces to assist the development of 
\begin{itemize}
    \item extensions for the language accepted by grounder \gringo (the functionality of  makes  the Pre-processor component required in the architecture of \ezcsp obsolete)
to accommodate, for example, the irregular atoms as discussed here; 
\item extensions for implementing specialized propagators to accommodate, for example, the processing of irregular atoms as discussed here natively and efficiently utilizing the fact that these propagators are defined within \clingo itself.
\end{itemize}

It is interesting to note that one can view/name constraint answer set programming  as ASP modulo constraints/theories (following the tradition of SMT). An interesting related paradigm to CASP is called ASP modulo acyclicity~\cite{bom16}. In this paradigm, specialized propagator is used to capture constraints specific to graph/tree problems. 
 \citeN{bom16} describe an integrational solver for ASP modulo acyclicity based on answer set solver {\sc clasp} and use Hamiltonian cycle problem as one of the benchmarks to showcase the system.
 
\paragraph{Briefing: Input Languages of CASP systems.}
In this paper we speak in some detail about the \ez language that is used for problem encodings to interface CASP systems  \ezcsp and \ezsmt. The other CASP tools such as \clingcon, \clingodl, or \clingolp introduce their own ASP-like dialects to state CAS programs with schematic variables. At the moment, the task of  transferring an encoding designed for one CASP system into an encoding meant for another CASP system requires a programmer experienced with dialects of these systems. An effort in spirit of the design of the standard ASP-Core-2 Language~\cite{cal19} (to interface ASP solvers) is now due for the case of CASP languages.

\section{Translational Approach via System \ezsmt}

The concluding part of Section~\ref{sec:cassmt} describes how  given a constraint answer set program  one can construct an SMT formula whose models capture its  answer sets. This construction relies on the concepts of completion and level ranking.
It is worth noting that 
Janhunen~\citeyear{jan06a} 
introduced   refined ``strong'' and ``strongly connected component (SCC)'' level rankings for normal logic programs (under a name of level numberings).
These refined versions of level ranking can be used to reduce  the size of translation from a  program to an SMT formula. Shen and Lierler~\citeyear{shen18} generalized these results to logic programs whose rules are of the form~\eqref{e:rule}. 
These ideas are also applicable within CAS programs and are utilized in the implementation of the CASP solver~\ezsmt~\cite{shen18a}. We now review the key features of the \ezsmt system and its building blocks to showcase  a translational approach.

In a nutshell system \ezsmt translates a given CAS program into an SMT formula and 
then utilizes an SMT solver as its search back-end to find models of the formula. Then, each model found in this way is mapped to an answer set of the given program. 
In addition to difference logic, ILA, and LA logics,
system  \ezsmt can use such SMT-logics as AUFLIRA and AUFNIRA~\cite{aufliraurl}.
Logic AUFLIRA
enables us to state linear constraints that may simultaneously  contain integer and real variables. Logic AUFNIRA permits nonlinear constraints, too.
As mentioned earlier, the \ezsmt system accepts programs in the \ez language.
This language is extended by several directives that allow users to specify a domain for a constraint variable.

	\begin{figure}
		\footnotesize
		\begin{center}
			\includegraphics[scale=1.5]{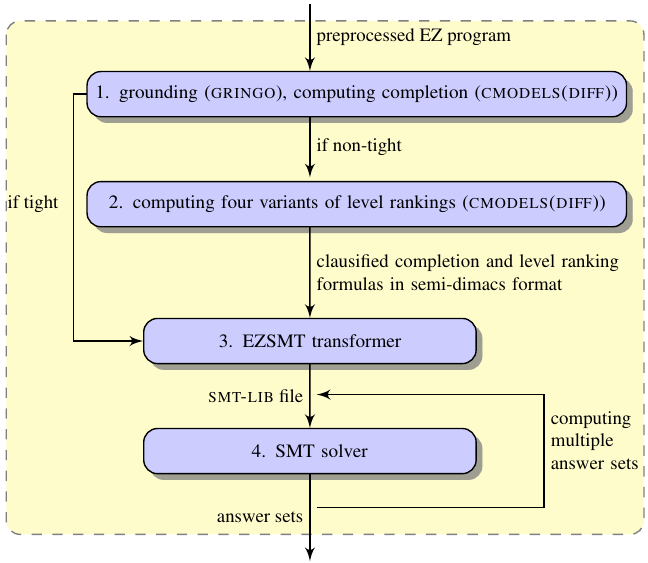}
			\normalsize
			\caption
			{System \ezsmt Architecture}
			\label{fig:ezsmtarch}
		\end{center}
	\end{figure}

Figure~\ref{fig:ezsmtarch} illustrates the  architecture of system \ezsmt. 
 The graphic is reproduced from the paper by Shen and Lierler~\shortcite{shen18a}. 
The system takes an EZ program as an input. It starts by applying the Pre-processor component of system \ezcsp (see Figure~\ref{fig:arch}); the rationale behind the application of this component is the same as in case of the \ezcsp system discussed in previous section.
It then utilizes grounder {\gringo}~\cite{Gebser2011} for eliminating ASP variables. 
Routines of system {\diff}~\cite{shen18} are used to compute input completion and level rankings of the program (Steps~1 and 2).
During Step 1, \ezsmt also determines  whether the program is ``tight'' or not. 
The tightness~\cite{fag94} is a syntactic condition on a program. Intuitively, a program is tight if it has no circular dependencies between its head and positive body atoms across a program. A simple example of a non-tight program is a program with a single rule $p\ar p$. In case when a program is tight it is sufficient to replace a formula in~\ref{ib2}  of Definition~\ref{def:tr1}
 by a simpler formula~\eqref{eq:comp2} stemming from the completion to achieve a one-to-one correspondence between the
 answer sets of a given program and the models of the corresponding SMT formula.
If the program is not tight, the corresponding level ranking formula is added.
A procedure used by  \ezsmt to perform this task  is identical to that of \diff~\cite{shen18}. System \ezsmt may construct different kinds of level ranking formulas including  strong level ranking formulas,  SCC level ranking formulas, and  strong SCC level ranking formulas, respectively. 
The resulting formulas are clausified to produce an output in semi-Dimacs format~\cite{sus16b} (Step 3.), which is 
 transformed
into \smtlib syntax --- a standard input language for SMT solvers~\cite{smt15} ---
using the procedure described by Susman and Lierler~\shortcite{sus16b} and Shen and Lierler~\shortcite{shen18a}.
Finally, one of the SMT solvers \cvcFour~\cite{cvc4}, \zThree~\cite{z3url}, or \yices~\cite{yicesurl} is called to compute models (Step 4.).
In fact, any other SMT solver supporting SMT-LIB can be utilized easily, too.
The \ezsmt system allows one to  compute multiple (extended) answer sets. It utilizes ideas exploited in the implementation of \diff~\cite[Section 5]{shen18}. In summary, after computing an (extended) answer set $X$ of a program \ezsmt invokes an SMT solver again by adding  formulas encoding the fact that a newly computed model should be different from~$X$. This process is repeated until the pre-specified number of solutions is enumerated or it has been established that no more solutions exist.
The described process of enumerating multiple solutions is naive and begs for an improvement. \citeN{geb07c} describe sophisticated methods for enumerating answer sets implemented within answer set solver \clasp.

        

\paragraph{Briefing: Translational Systems.}
We mentioned such translational constraint answer set solvers as~{\mingo}~\cite{liu12}, {\dingo}~\cite{jan11}, and  {\sc aspartame}~\cite{ban15}.
  To process CAS programs with LA and ILA logics, the~{\mingo} system computes program's input completion extended with level ranking formulas and then translates these formulas into mixed integer programming expressions. After that it uses the  {\sc cplex} solver~\cite{cplex}
to solve these formulas.
To process CAS programs with difference logic, system {\dingo}  translates these programs into  SMT(DL) formulas using translations in spirit of those in  \ezsmt and applies the SMT solver {\sc z3}~\cite{z3} to find their models.
The last translational system that we mention is  \aspmt~\cite{Bartholomew2014}.  
The \aspmt system is a close relative of \ezsmt in the sense that it utilizes SMT solver \zThree for search. Solver \aspmt is nevertheless restricted to tight programs. It computes the completion of a given program and then invokes \zThree solver to enumerate the solutions.
System {\sc aspartame} differs from all of the above as it translates the CAS programs with IL arithmetic into answer set programs.

\section{Big Picture and Experimental Data}

We start this section by summarizing the modern landscape of  CASP technology. We then proceed towards presenting some experimental data to showcase the current computational capabilities of the field.

Figure~\ref{fig:table} reproduces part of the table stemming from~Janhunen et al.  \citeyear{jan17} that provides a great overview of the key features and capabilities of the constraint answer set programming systems (the only difference between the original table and the one present is that system \ezsmt is now marked as the one capable of processing non-tight programs; that was not the case prior). 
In the first row of the table, {\em integrational} is taken to be the complement of {\em translational}. 
The row {\em real numbers} refers to the ability of a solver to support constraints over real numbers. Every solver supports constraints over integers. The row {\em optimization} refers to the ability of a solver to support optimization statements that are valuable in designing solutions to real world problems~\cite{and12}.  
The row {\em non-tight} points to systems that do not require input programs to satisfy a syntactic condition of tightness.  
The table  introduces two more systems {\sc inca}~\cite{dre11a,dre11} and {\sc dlvhex[cp]}~\cite{ros15}, which have not surfaced in our discussion. System {\sc inca} implements a lazy propagation-translation approach that, along processing time, translates integer constraints of the program into logic program rules. 
System {\sc dlvhex[cp]} is a {\sc clingo}-based system that utilizes CSP solver {\sc gecode}. 

\begin{figure}
\scriptsize
\begin{tabular}{cccccccccccc}
\hline
&\clingo&\clingo&{\sc cling}&{\sc aspar}      &{\sc inca}&{\sc ez}& {\sc ez}&\mingo&\dingo&{\sc aspmt}&{\sc dlvhex}\\
&{\sc [dl]}&{\sc [lp]}&{\sc con}  &{\sc tame} &          &{\sc csp} & {\sc smt} &      &      &{\sc 2smt} &{\sc [cp]}\\
\hline
translational&\xmark&\xmark&\xmark&\cmark&\xmark&\xmark&\cmark&\cmark&\cmark&\cmark&\xmark\\
real numbers &\xmark&\cmark&\xmark&\xmark&\xmark&\cmark&\cmark&\cmark&\xmark&\cmark&\xmark\\
optimization &\xmark&\cmark&\cmark&\cmark&\cmark&\xmark&\xmark&\xmark&\xmark&\xmark&\cmark\\
non-tight    &\cmark&\cmark&\cmark&\cmark&\cmark&\cmark&\cmark&\cmark&\cmark&\xmark&\cmark\\
\end{tabular}
     \normalsize
     \caption{Solvers and their features\label{fig:table}}
\end{figure}

The experimental data presented here is reproduced from the paper by Shen and Lierler~\shortcite{shen18a}.
It focuses on the performance of three systems
 \ezsmt, \ezcsp and \clingcon~\cite{ost17,ost18}.
The benchmarks are posted at the \ezsmt website (see Footnote~\ref{foot:ezsmt}).
In conclusion of this section we remark on how these systems compare to other CASP solvers.

We now  point to the origins of the considered benchmarks.
Three benchmarks, namely, Reverse Folding (RF), Incremental Scheduling (IS), and Weighted Sequence (WS), come from the Third Answer Set Programming Competition~\cite{aspcomp3}. We obtain \hbox{\clingcon} and \ezsmt encodings of IS from Banbara et al. \citeyear{ost17}. We include a benchmark problem called Blending (BL)~\cite{sar17} and extend it to BL*, which contains variables over both integers and reals.
Also, we use the Bouncing Ball (BB) domain~\cite{micphd}. It is important to remark that
the  encoding for BB domain results in a tight program.
This domain uses nonlinear constraints over real numbers.
Three more benchmarks, namely, RoutingMin (RMin), RoutingMax (RMax), and Travelling Salesperson (TS) are obtained from~Liu et al.~\citeyear{liu12}. 
The obtained TS benchmark is an optimization problem that we turn into a TS variant considered in the introduction.
The  Labyrinth (LB) benchmark is extended from the domain presented in the Fifth Answer Set Programming Competition~\cite{aspcomp5}.
This extension allows us to add  integer linear constraints into the problem encoding.
The next benchmark, Robotics (RB), comes from Young et al.~\citeyear{you17}.
Also, we present results on two benchmarks from Balduccini et al.~\citeyear{bal17}, namely, Car and Generator~(GN).

All benchmarks are run on an Ubuntu 16.04.1 LTS (64-bit) system with an Intel core i5-4250U processor. The resource allocated for each benchmark is limited to one CPU core and 4GB RAM. We set a timeout of 1800 seconds. No problems are solved simultaneously.
The systems that we use to compare the performance of variants of
\ezsmt (invoking SMT solver
\zThree v.~4.5.1;
\yices  v.~2.5.4) are
\clingcon  v.~3.3.0 and the variants of
\ezcsp v.~2.0.0 (invoking ASP solvers \clasp~v.~3.2.0; and 
constraint  solver 
\sp v.~7.4.1 or  \mzn~v.~2.0.2).
The \gringo system v.~4.5.3 is used as grounder for \ezsmt and \ezcsp with one exception: \gringo v.~3.0.5 is utilized for \ezcsp for the Reverse Folding benchmark (due to some incompatibility issues).

\begin{figure}[h]
\scriptsize
\centering
\begin{tabular}{c|c|c|cccc|cc}
  \hline
Category & {Benchmark} & \clingcon & \multicolumn{2}{c}{{\ezsmt}(\zThree)} & \multicolumn{2}{c|}{{\ezsmt}(\yices)} & \ezcsp  &  \ezcsp  \\
 & & & SCC & SCCStrong & SCC & SCCStrong & \clasp-\swi & \clasp-\mzna\\
  \hline
{NT-IL} & RMin (100)&  {\bf 4.68} & 8.76 & 11.8& 5.81 &7.57& 126007(70)& 126002(70)\\
\hline{~--------}
 & RMax (100)&  3144 & 459 & {\bf 22.4}& 5190 & 5945& 180000(100)& 180000(100)\\
\hline{~--------}
 & TS (30)& 455 & 7347(4) & 43620(24)& 1881(1)  & 75.2& {\bf 14.3}& 54000(30)\\
\hline{~--------}
 & LB (22)& {\bf 3002(1)} & 9510(1) & 10089(2)& 4399(2) & 5512(2)& 12558(6)& 12638(6)\\
\hline
\hline
{T-IL} & RF (50)& 326 &  \multicolumn{2}{c}{6058(2)} &  \multicolumn{2}{c|}{27840(14)} & {\bf 101}& 7218(4)\\
\hline{~--------}
 & IS (30)& {\bf 9080(5)} & \multicolumn{2}{c}{ 9200(5) }& \multicolumn{2}{c|}{  9098 (5)}&  41446(21) & 39458(21)\\
\hline{~--------}
 & WS (30)& 52.5 & \multicolumn{2}{c}{ 29.2}& \multicolumn{2}{c|}{ {\bf 5.23}}&  54000(30)& 54000(30)\\
\hline
\hline
T-INL & Car (8)& / & \multicolumn{2}{c}{0.32} & \multicolumn{2}{c|}{{\bf 0.25}} & 10.1 & 2.34 \\
\hline
\hline
{T-RL} & BL (30)& / & \multicolumn{2}{c}{ 88.4} & \multicolumn{2}{c|}{ {\bf 47.4} }& 18322(9)& / \\
\hline{~--------}
 & GN (8)& / & \multicolumn{2}{c}{0.58} & \multicolumn{2}{c|}{{\bf 0.48}}& 5641(3) & / \\
\hline{~--------}
 & RB (8)& / & \multicolumn{2}{c}{0.4} & \multicolumn{2}{c|}{{ \bf 0.39}} & 2.04 & / \\
\hline
\hline
 {T-RNL} & BB (5)& / & \multicolumn{2}{c}{ 3663(2)}& \multicolumn{2}{c|}{ {\bf 0.98}}& 9000(5)& / \\
\hline
\hline
T-ML & BL* (30)& / & \multicolumn{2}{c}{ {\bf 5573(2)}} & \multicolumn{2}{c|}{ / }& / & /\\
\hline
\hline
\end{tabular}
\normalsize

\caption{Experimental Data\label{fig:experiments}}
\end{figure}
\begin{figure}[h!]
\small
\begin{tabular}{c|c||c|c|c||c|c}
\hline
T     &NT     &I       & R    & M     &L      &NL     \\
\hline
tight&non-tight&integer & real & mixed &linear&non linear\\
\hline
\end{tabular}
\normalsize
\caption{Meaning of the Category Column Letters\label{fig:letters}}
\end{figure}

Figure~\ref{fig:experiments} summarizes the main  results. In this figure,
we use {\ezsmt}(\zThree) and {\ezsmt}(\yices) to denote two variants of \ezsmt.
Acronym \ezcsp-\clasp-\swi \hbox{(\ezcsp-\clasp-\mzna)} stands for a  variant of \ezcsp, where \clasp is utilized as the answer set solver and \sp (\mzn, respectively) is utilized as a constraint solver.

In Figure~\ref{fig:experiments}, 
we present cumulative time in seconds of all instances for each benchmark with numbers of unsolved instances due to timeout or insufficient memory inside parentheses. The  ``/" sign indicates that this solver or its variant does not support the kinds of constraints occurring in the encoding. For example, \clingcon does not support constraints over real numbers or nonlinear constraints.
The total number of used instances is shown in parentheses after a benchmark name.
All the steps involved, including grounding and transformation, are reported as parts of the solving time. 
The benchmarks are divided into categories by double separations. 
Figure~\ref{fig:letters} presents the readings of the letters in the category column, where the first two letters refer to the syntactic condition on a logic program; the middle three letters refer to the domains of constraint variables of the program; and the last two letters refer to the kinds of constraints.


Systems \clingcon, \ezcsp-\clasp-\swi, and \hbox{\ezcsp-\clasp-\mzna}
are run in their default settings.   For non-tight programs,
system \ezsmt with strongly connected components level rankings (flags {\tt -SCClevelRanking} and {\tt -SCClevelRankingStrong}) show best performance.

In summary, we observe that \clingcon achieves first positions in three benchmarks. \ezcsp-\clasp-\swi and {\ezsmt}(\zThree) win in two benchmarks, respectively. {\ezsmt}(\yices) ranks first in six benchmarks.
The {\ezsmt}(\yices) system displays the best overall results.
Utilizing different SMT solvers may improve the performance of \ezsmt in the future.


\def \cldl{{\sc clingo[dl]}\xspace}
\def \cllp{{\sc clingo[lp]}\xspace}

\paragraph{On anticipated performance of related systems.}
CASP solver \cllp~\cite{jan17}  handles   linear constraints over integers or reals. 
The experimental analysis presented by Janhunen et al.~\shortcite{jan17} only considers programs with constraints over integers. On these benchmarks, 
\clingcon outperforms \cllp.  
Susman and Lierler~\shortcite{sus16b} compare the performance of \mingo~\cite{liu12} and \ezsmt on tight programs. The latter consistently has better performance. 
The \aspmt~\cite{Bartholomew2014} system is a close relative of \ezsmt in the sense that it utilizes SMT solver \zThree for search. 
  We expect that {\ezsmt}(\zThree) times mimic these of \aspmt on tight programs.

\section{Discussion and Future Directions}
This article was meant to construct a compelling tale of constraint answer set programming developments of the past decade supplying the interested reader with birds-eye view of the area and enough literature links to acquire details when needs be. This concluding section lists open questions and possible directions of the field.

Gebser et al.~\shortcite{geb16} point out how an ASP-based problem solving  frequently requires capabilities going  beyond classical ASP language and systems. They observe that ASP system \clingo and/or its grounder {\gringo} and/or its solver \clasp often serve as important building blocks of more complex systems (including such systems as constraint answer set solvers). The fundamental contribution by Gebser et al.~\shortcite{geb16} was to conceive \clingo~5 framework that provides a general purpose interface which  helps to make extensions of \gringo/\clasp systems a routine and systematic process. This interface also targets facilitation of streamlining communication between theory/constraint propagation and answer set solving propagation as well as other advanced techniques such as conflict driven learning implemented in \clasp. 
As such \clingo~5 can be seen as one of the key contributions to the CASP community.  It provides a general purpose platform for bootstrapping unique  constraint answer set programming solutions.

Automated reasoning spans areas such as satisfiability solving, answer set programming, satisfiability modulo theories solving, integer (mixed) programming, constraint answer set programming, and constraint processing. The relation between answer set solvers and  satisfiability solvers is well understood, see, for example, the paper by~\citeN{lier16a}. Also, the relation between different instances of answer set solvers has been studied: see the paper by Lierler and Truszczynski~\shortcite{lier11}.
Several representatives of the integration approach to constraint answer programming have been contrasted and compared, see, for instance, the paper by~\citeN{lier14}.
Yet,  a deeper understanding of how various solving techniques in theory solving of SMT compare to these of constraint processing CSP/CLP is missing. 
Similarly, the following is an open question: how do techniques in mixed (integer) programming compare to these in SMT, CSP, and/or integer linear programming. At the moment the best we can do is to use constraint answer set programming and its various implementations that include translational approaches to conduct experimental analysis that spans a variety of automated reasoning communities~\cite{lie17,jan17}. 
\citeN{dov09} provide us with insights on how CLP solutions to combinatorial search problems compare to these with ASP solutions.
Deeper understanding of differences and similarities between algorithms used in these traditionally different areas of AI is due.

Papers by~\citeN{sel03}, \citeN{lif17}, \citeN{fan20}, Cabalar et al.~\shortcite{cab20b},~\citeN{bom20}, to name a few, provide the techniques for analysing and arguing program correctness in traditional answer set programming. To the best of our knowledge there were no attempts to lift these methods to the scope of constraint answer set programming. As hybrid answer set programming approaches are making their pronounced way into practice the methodologies for arguing the correctness of such solutions are due. 
Alternative definitions of CAS programs and their semantics as studied by \citeN{lin08},  \citeN{bal13a}, and ~\citeN{bar13} may suit the purpose of the formal arguments of correctness better than the definition presented here.

One way to view translational methods in constraint answer set programming is as an attempt to utilize existing technology from distinct automated reasoning subfields for solving CASP formulations of solutions to problems. Another way to view these is as an attempt to provide a  programming front end of logic programming under answer set semantics to the variety of tools that otherwise possess only limited modeling capabilities. For example, despite the existence of the common standard SMT-LIB language for formulating SMT problems one may not call that a full-fledged programming language. Just as the DIMACS format -- standard for communicating with the satisfiability solvers -- does not constitute a suitable language for modeling solutions to problems in it directly. As mentioned earlier, \clingo~5 framework provides infrastructure for bootstrapping novel hybrid answer set programming solutions. It remains to be seen if this framework is sufficient for establishing a full-fledged front end for the translational approach targeting the utilization of SMT solvers that goes beyond traditional (integer) linear arithmetic. For example, particular SMT fragments provide vector and array arithmetics. It is still to be established whether expressions of these logics  may prove to be convenient modeling tools (backed up by specialized efficient search techniques of SMT).

MiniZinc\footnote{https://www.minizinc.org/} is a free and open-source constraint modeling language. In the past decade it became a standard front end for accessing a conglomerate of  CSP tools. The translational approaches of constraint answer set solvers  looked into utilizing SMT solvers via SMT-LIB so far. In a similar manner, the language of MiniZinc can be utilized for accessing CSP tools that support the MiniZinc language. MiniZinc-based translational approach to constraint answer set solving is still to see the light.

At the closing of Section~\ref{sec:int}, we  mention that  no efforts by the research community have been taken to produce a standard input language for CASP solvers. The maturity of the field suggests it is time for such an effort. Possibly, an even more ambitious effort is due. This paper makes it clear that many automated reasoning paradigms --- SMT, ASP, CASP, CSP, CLP -- are geared towards solving difficult combinatorial search problems. We named several case studies, where researchers attempt to experimentally compare these methods by designing solutions to problems in distinct paradigms and then studying behaviors of respective solvers on these solutions. Providing a standard language to interface tools from distinct communities will allow us   to benefit from portfolio approaches~\cite{nud04}  originated in SAT by tapping into a broad spectrum of solving techniques. We trust that the standard language for CASP together with translational techniques that are able to transform CAS programs into the specifications in languages of related paradigms is a promising directions of research. To this end the question of a mature programming methodology for utilizing CASP is in need. At the moment, typical users of ASP are the ones that practice CASP. They borrow so called generate define and test methodology of ASP~\cite{lif02,den19} that accounts for logic programming aspect of CASP. Yet, the methodology that naturally accounts for constraints is still to come.     

Optimization statements are important  in (constraint) answer set programming. As one can see in Figure~\ref{fig:table} no translational approach supports these statements. Utilizing MiniZinc/CSP solvers would allow to elevate this restriction as CSP solvers typically provide support for optimization problems. Also, MaxSMT~\cite{rob10} concerns  SMT solving that provides means to formulate optimization statements. It is yet another direction of work on connecting MaxSMT together with optimization statements of constraint answer set programming.

\section*{Acknowledgments}
I would like to acknowledge and cordially thank many of my collaborators with whom we had a chance to contribute to an exciting field of Constraint Answer Set Programming and many of my colleagues who have fostered  my understanding of the subject matter:  Marcello Balduccini, Broes De Cat, Marc Denecker, Martin Gebser, Michael Gelfond, Tomi Janhunen, Roland Kaminski, Martin Nyx Brain, Joohyung Lee, Ilkka Niemel\"a, Max Ostrowski, Torsten Schaub, Peter Schueller, Da Shen, Benjamin Susman, Cesare Tinelli, Miroslaw Truszczynski,  Philipp Wanko, Yuanlin Zhang. Thank you for the years of an incredible journey. Also, I would like to thank the anonymous reviewers for their valuable feedback, which helped to bring the article to this form.
This work was partially supported by the NSF 1707371 grant.
\bibliographystyle{acmtrans}
 \bibliography{shared}
 
\label{lastpage}
 
\end{document}